\documentclass[11pt,a4paper]{article}
\usepackage[hyperref]{udst2018}
\usepackage{latexsym}
\usepackage{graphics}
\usepackage{tikz}
\usepackage{todonotes}
\usepackage{amsmath}
\usepackage{xcolor}
\usepackage{colortbl}
\usepackage{url}
\usepackage{enumitem}
\usepackage[normalem]{ulem}
\usepackage{setspace}
\usepackage{lmodern}
\usepackage{slantsc}
\usepackage[T1]{fontenc}
\usepackage[utf8]{inputenc}
\usepackage[icelandic,english]{babel}
\usepackage{times}

\aclfinalcopy 
\def\aclpaperid{***} 

\title{82 Treebanks, 34 Models: Universal Dependency Parsing with Multi-Treebank Models}

\author{Aaron Smith$^{*}$~~~Bernd Bohnet$^{\dagger}$~~~Miryam de Lhoneux$^{*}$\\[1mm] 
\textbf{Joakim Nivre}$^{*}$~~~\textbf{Yan Shao}$^{*}$~~~\textbf{Sara Stymne}$^{*}$ \\[3mm]
\begin{tabular}{cp{10mm}c}
$^{*}$Department of Linguistics and Philology && $^{\dagger}$Google Research \\ 
Uppsala University && London, UK \\ 
Uppsala, Sweden && \\
  \end{tabular}
}

\date{}

\begin{document}
\maketitle

\begin{abstract}
We present the Uppsala system for the CoNLL 2018 Shared Task on 
universal dependency parsing.
Our system is a pipeline consisting of three components: the first performs joint word and sentence segmentation; the second predicts part-of-speech tags and morphological features; the third predicts dependency trees from words and tags. 
Instead of training a single parsing model for each treebank, we trained models with multiple treebanks for 
one language or closely related languages, greatly reducing the number of models.
On the official test run, we ranked 7th of 27 teams for the LAS and MLAS metrics.

Our system obtained the best scores overall for word segmentation, universal POS tagging, and morphological features.
\end{abstract}

\section{Introduction}
\label{sec:intro}

The CoNLL 2018 Shared Task on Multilingual Parsing from Raw Text to Universal Dependencies \citep{conll2018} requires participants to build systems that take as input raw text, without any linguistic annotation, and output full labelled dependency trees for 82 test treebanks covering 46 different languages.
Besides the labeled attachment score (LAS) used to evaluate systems in the 2017 edition of the Shared Task \citep{udst:overview}, this year's task introduces two new metrics: morphology-aware labeled attachment score (MLAS) and bi-lexical dependency score (BLEX).
The Uppsala system focuses exclusively on LAS and MLAS,
and consists of a three-step pipeline.
The first step is a model for joint sentence and word segmentation which uses the BiRNN-CRF framework of \citet{shao17,shao18} to predict sentence and word boundaries in the raw input and simultaneously marks multiword tokens that need non-segmental analysis. 
The second component is a part-of-speech (POS) tagger based on \citet{bohnet18acl}, which employs a sentence-based character model and also predicts morphological features.
The final stage is a greedy transition-based dependency parser that takes segmented words and their predicted POS tags as input and produces full dependency trees. 
While the segmenter and tagger models are trained on a single treebank, the parser uses multi-treebank learning to boost performance and 
reduce the number of models.

After evaluation on the official test sets \citep{ud22data}, which was run on the TIRA server \citep{tira}, the Uppsala system ranked 7th of 27 systems with respect to LAS, with a macro-average F1 of 72.37, and 7th of 27 systems with respect to MLAS, with a macro-average F1 of 59.20.
It also reached the highest average score for word segmentation (98.18), universal POS (UPOS) tagging (90.91), and morphological features (87.59).

\paragraph{Corrigendum:} After the test phase was over, 
we discovered that we had used a non-permitted resource when developing the UPOS tagger for Thai PUD (see Section
\ref{sec:tagger}). Setting our LAS, MLAS and UPOS scores to 0.00 for Thai PUD gives the corrected scores: LAS 72.31, MLAS 59.17, UPOS 90.50. This does not affect the ranking for any of the three scores, as confirmed by the shared task organizers.

\section{Resources}
\label{sec:resources}
All three components of our system were trained principally on the training sets of Universal Dependencies v2.2 released to coincide with the shared task \citep{ud22data}. 
The tagger and parser also make use of the pre-trained word embeddings provided by the organisers, as well as Facebook word embeddings \cite{bojanowski2016enriching}, and both word and character embeddings trained on Wikipedia text\footnote{\url{https://dumps.wikimedia.org/backup-index-bydb.html}} with word2vec \citep{mikolov2013distributed}.
For languages with no training data, we also used external resources in the form of Wikipedia text, parallel data from OPUS \citep{tiedemann2012OPUS}, the Moses statistical machine translation system \citep{Moses07}, and the Apertium morphological transducer for Breton.\footnote{\url{https://github.com/apertium/apertium-bre}}

\section{Sentence and Word Segmentation}
\label{sec:segmentation}

We employ the model of \citet{shao18} for joint sentence segmentation and word segmentation. 
Given the input character sequence, we model the prediction of word boundary tags as a sequence labelling problem using a BiRNN-CRF framework \citep{huang2015bidirectional,shao17}.
This is complemented with an attention-based LSTM model \citep{bahdanau2014neural} for transducing non-segmental multiword tokens. 
To enable joint sentence segmentation, we add extra boundary tags as in \citet{uu-conll17}.

We use the default parameter settings introduced by \citet{shao18} and train a segmentation model for all treebanks with at least 50 sentences of training data. 
For treebanks with less or no training data (except Thai discussed below), we substitute a model for another treebank/language:
\begin{itemize}[topsep=2pt,itemsep=1pt]
\item For Japanese Modern, Czech PUD, English PUD and Swedish PUD, we use the 
model trained on the largest treebank from the same language (Japanese GSD, Czech PDT, English EWT and Swedish Talbanken
).
\item For Finnish PUD, we use Finnish TDT rather than the slightly larger Finnish FTB, because the latter does not contain raw text suitable for training a segmenter. 
\item For Naija NSC, we use English EWT. 
\item For 
other test sets with little or no training data, we select 
models based on the size of the intersection of the character sets measured on Wikipedia data (see Table \ref{tab:results} for details).\footnote{North Sami Giella was included in this group by mistake, as we underestimated the size of the treebank.} 
\end{itemize}

\paragraph{Thai}
Segmentation of Thai was a particularly difficult case: Thai uses a unique script, with no spaces between words, and there was no training data available.
Spaces in Thai text can function as sentence boundaries, but are also used equivalently to commas in English
.
For Thai sentence segmentation, we 
exploited the fact that four other datasets are parallel, i.e., there is a one-to-one correspondence between sentences in Thai and in Czech PUD, English PUD, Finnish PUD and Swedish PUD.\footnote{This information was available in the README files distributed with the training data and available to all participants.}
First, we split the Thai text by white space and treat the obtained character strings as potential sentences or sub-sentences.
We then align them to the segmented sentences of the 
four parallel datasets using the Gale-Church algorithm \citep{gale1993program}. 
Finally, we compare the sentence boundaries obtained from different parallel datasets and adopt the ones that are shared within at least three parallel datasets. 

For word segmentation, we use a trie-based segmenter with a word list derived from the Facebook word embeddings.\footnote{\url{github.com/facebookresearch/fastText}} The segmenter retrieves words by greedy forward maximum matching \citep{wong96}.  
This method requires no training but gave us the highest word segmentation score of 69.93\% for Thai, compared to the baseline score of 8.56\%.

\section{Tagging and Morphological Analysis}
\label{sec:tagger}

We use two separate instantiations of the tagger\footnote{\url{https://github.com/google/meta_tagger}} described in \citet{bohnet18acl} to predict UPOS tags and morphological features, respectively. The tagger uses a Meta-BiLSTM over the output of a sentence-based character model and a word model. 
There are two features that mainly distinguishes the tagger from previous work. The character BiLSTMs use the full context of the sentence in contrast to most other taggers which use words only as context for the character model. This character model is combined with the word model in the Meta-BiLSTM relatively late, after two layers of BiLSTMs.

For both the word and character models, we use two layers of BiLSTMs with 300 LSTM cells per layer. We employ batches with 8000 words and 20000 characters. We keep all other hyper-parameters as defined in \newcite{bohnet18acl}. From the training schema described in the above paper, we deviate slightly in that we perform early stopping on the word, character and meta-model independently. We apply early stopping due to the performance of the development set (or training set when no development set is available) and stop when no improvement occurs in 1000 training steps. We use the same settings for UPOS tagging and 
morphological features. 

To deal with languages that have little or no training data, we adopt three different strategies:
\begin{itemize}[topsep=2pt,itemsep=1pt]
\item For the PUD treebanks (except Thai), Japanese Modern and Naija NSC, we use the same model substitutions as for segmentation (see Table \ref{tab:results}). 
\item For Faroese we used the model for Norwegian Nynorsk, as we believe this to be the most closely related language. 
\item For treebanks with small training sets 
we use only the provided training sets for training. 
Since these treebanks do not have development sets, we use the training sets for early stopping as well.
\item For Breton and Thai, which have no training sets and no suitable substitution models, we use a bootstrapping approach to train taggers as described below.
\end{itemize}

\paragraph{Bootstrapping} 
We first annotate an unlabeled corpus using an external morphological analyzer. We then create a (fuzzy and context-independent) mapping from the morphological analysis to universal POS tags and features, which allows us to relabel the annotated corpus and train taggers using the same settings as for other languages
. For Breton, we annotated about 60,000 sentences from Breton OfisPublik, which is part of OPUS,\footnote{\url{https://opus.nlpl.eu/OfisPublik.php}} using the Apertium morphological analyzer. The Apertium tags could be mapped to universal POS tags and a few morphological features like person, number and gender. For Thai, we annotated about 33,000 sentences from Wikipedia using PyThaiNLP\footnote{\url{https://github.com/PyThaiNLP/pythainlp/wiki/PyThaiNLP-1.4}} and mapped only to UPOS tags (no features). Unfortunately, we realized only after the test phase that PyThaiNLP was not a permitted resource
, which invalidates our UPOS tagging scores for Thai, as well as the LAS and MLAS scores which depend on the tagger. Note, however, that the score for morphological features is not affected, as we did not predict features at all for Thai. The same goes for sentence and word segmentation, which do not depend on the tagger.

\paragraph{Lemmas}
Due to time constraints we chose not to focus on the BLEX metric in this shared task.
In order to avoid zero scores, however, we simply 
copied a lowercased version of the raw token into the lemma column.

\section{Dependency Parsing}
\label{sec:parsing}

We use a greedy transition-based parser \citep{nivre2008algorithms} based on the framework of \citet{kiperwasser16} where
BiLSTMs \citep{hochreiter1997long,graves2008bilstms} learn representations of tokens in context, and are trained together with a multi-layer perceptron that predicts transitions and arc labels based on a few Bi\-LSTM vectors.
allow the construction of non-projective dependency trees \cite{nivre09acl}.
We also introduce a static-dynamic oracle to allow the parser to learn from non-optimal configurations at training time in order to recover better from mistakes at test time \citep{delhoneux17arc}.

In our parser, the vector representation $x_i$ of a word type $w_i$ before it is passed to the BiLSTM feature extractors is given by:
\vspace{-1.5mm}
\begin{equation*}
    x_i = e(w_i) \circ e(p_i) \circ \text{BiLSTM}(ch_{1:m}) .
\vspace{-1.5mm}
\end{equation*} 
Here, $e(w_i)$ represents the word embedding and $e(p_i)$ the POS tag embedding \citep{chen14}; these are concatenated to a character-based vector, obtained by running a BiLSTM over the characters $ch_{1:m}$ of $w_i$.

With the aim of training multi-treebank models, we additionally created a variant of the parser which adds a treebank embedding $e(tb_i)$ to input vectors in a spirit similar to the language embeddings of \citet{ammar16} and \citet{uu-conll17}: 
\vspace{-1.5mm}
\begin{equation*}
    x_i = e(w_i) \circ e(p_i) \circ \text{BiLSTM}(ch_{1:m}) \circ e(tb_i).
\vspace{-1.5mm}
\end{equation*}
an effective way to combine multiple monolingual heterogeneous treebanks \citep{stymne+18acl} and applied them to low-resource languages \citep{uu-conll17}.
In this shared task, the treebank embedding model was used both monolingually, to combine several treebanks for a single language, and multilingually, mainly for closely related languages, both for languages with no or small treebanks, and for languages with medium and large treebanks, as
described in Section~\ref{sec:multilingual}.

During training, a word embedding for each word type in the training data is initialized using the pre-trained embeddings provided by the organizers where available.
For the remaining languages, we use different strategies:
\begin{itemize}[topsep=2pt,itemsep=1pt]
\item For Afrikaans, Armenian, Buryat, Gothic, Kurmanji, North Sami, Serbian and Upper Sorbian, we carry out our own pre-training on the Wikipedia dumps of these languages, tokenising them with the baseline UDPipe models and running the implementation of word2vec in the Gensim Python library\footnote{\url{https://radimrehurek.com/gensim/}} with 30 iterations and a minimum count of 1. 
\item For Breton and Thai, we use specially-trained multilingual embeddings (see Section~\ref{sec:multilingual}). 
\item For Naija and Old French, we substitute English and French embeddings, respectively.
\item For Faroese, we do not use pre-trained embeddings. While it is possible to train such embeddings on Wikipedia data, as there is no UD training data for Faroese we choose instead to rely on its similarity to other Scandinavian languages (see Section~\ref{sec:multilingual}).
\end{itemize}
Word types in the training data that are not found amongst the pre-trained embeddings are initialized randomly using Glorot initialization \cite{glorot2010understanding}, as are all POS tag and treebank embeddings.
Character vectors are also initialized randomly, except for Chinese, Japanese and Korean, in which case we pre-train character vectors using word2vec on the Wikipedia dumps of these languages.
At test time, we first look for out-of-vocabulary (OOV) words and characters (i.e., those that are not found in the treebank training data) amongst the pre-trained embeddings and otherwise assign them a trained OOV vector.\footnote{An alternative strategy is to have the parser store embeddings for all words that appear in either the training data or pre-trained embeddings, but this uses far more memory.}
A variant of word dropout is applied to the word embeddings, as described in \citet{kiperwasser16b}, and we apply dropout also to the character vectors. 

plus first item on the buffer with its leftmost dependent). 
We train all models for 30 epochs with hyper-parameter settings shown in Table~\ref{tbl:hyper}.
Note our unusually large character embedding sizes; we have previously found these to be effective, especially for morphologically rich languages \citep{smith18}. 
Our code is publicly available. We release the version used here as UUParser 2.3.\footnote{\url{https://github.com/UppsalaNLP/uuparser}}

\begin{table}[t]
    \begin{center}
        \begin{scalebox}{0.85}{
            \begin{tabular}{ l | c  }
                \hline 
                Character embedding dimension & 500 \\
                Character BiLSTM layers & 1 \\
                Character BiLSTM output dimension & 200 \\
                \hline
                Word embedding dimension & 100 \\
                POS embedding dimension & 20 \\
                Treebank embedding dimension & 12 \\
                Word BiLSTM layers & 2 \\
                Word BiLSTM hidden/output dimension & 250 \\
                \hline
                Hidden units in MLP & 100 \\  
                \hline
                Word dropout & 0.33 \\
                $\alpha$ (for OOV vector training) & 0.25 \\
                Character dropout & 0.33 \\
                $p_{agg}$ (for exploration training) & 0.1 \\
                \hline
            \end{tabular}}
        \end{scalebox}
        \caption{Hyper-parameter values for parsing.}
        \label{tbl:hyper}
    \end{center}
\end{table}

\paragraph*{Using Morphological Features}

\indent Having a strong morphological analyzer, we were interested in finding out whether or not we can improve parsing accuracy using predicted morphological information. We conducted several experiments on the development sets for a subset of treebanks. However, no experiment gave us any improvement in terms of LAS and we decided not to use this technique for the shared task. 

What we tried was to create an embedding representing either the full set of morphological features or a subset of potentially useful features, for example case (which has been shown to be useful for parsing by \citet{kapociute13lithuanian} and \citet{eryigit08cl}), verb form and a few others. That embedding was concatenated to the word embedding at the input of the BiLSTM. We varied the embedding size (10, 20, 30, 40), tried different subsets of morphological features, and tried with and without using dropout on that embedding. We also tried creating an embedding of a concatenation of the universal POS tag and the Case feature and replace the POS embedding with this one. We are currently unsure why none of these experiments were successful and plan to investigate this in the future. It would be interesting to find out whether or not this information is captured somewhere else. A way to test this would be to use diagnostic classifiers on vector representations, as is done for example in \citet{hupkes2018visualisation} or in \citet{adi17fine}.

\section{Multi-Treebank Models}
\label{sec:multilingual}

One of our main goals was to leverage information across treebanks to improve performance and reduce the number of parsing models. 
We use two different types of models:
\begin{enumerate}[topsep=2pt,itemsep=1pt]
	\item Single models, where we train one model per treebank (17 models applied to 18 treebanks, including special models for Breton KEB and Thai PUD).
    \item Multi-treebank models
    \begin{itemize}[topsep=0pt,itemsep=1pt]
     \item Monolingual models, 
    based on multiple treebanks for one language (4 models, trained on 10 treebanks, applied to 11 treebanks).
    \item Multilingual models, based on 
    treebanks from several (mostly) closely related languages (12 models, trained on 48 treebanks, applied to 52 treebanks; plus a special model for Naija NSC).
    \end{itemize} 
\end{enumerate}
When a multi-treebank model is applied to a test set from a treebank with training data, we naturally use the treebank embedding of that treebank also for the test sentences. However, when parsing a test set with no corresponding training data, we have to use one of the other treebank embeddings. In the following, we refer to the treebank selected for this purpose as the \emph{proxy} treebank (or simply \emph{proxy}).

In order to keep the training times and language balance in each model reasonable, we cap the number of sentences used from each treebank to 15,000, with a new random sample selected at each epoch. 
This only affects a small number of treebanks, since most training sets are smaller than 15,000 sentences.
For all our multi-treebank models, we apply the treebank embeddings described in Section \ref{sec:parsing}.
Where two or more treebanks in a multilingual model come from the same language, we use separate treebank embeddings for each of them. 
We have previously shown that multi-treebank models can boost LAS in many cases, especially for small treebanks, when applied monolingually \cite{stymne+18acl}, and applied it to low-resource languages \cite{uu-conll17}. 
In this paper, we add POS tags and pre-trained embeddings to that framework, and extend it to also cover multilingual parsing for languages with varying amounts of training data.
 
Treebanks sharing a single model are grouped together in Table \ref{tab:results}.
To decide which languages to combine in our multilingual models, we use two sources: knowledge about language families and language relatedness, and clusterings of treebank embeddings from training our parser with all available languages.
We created clusterings by training single parser models with treebank embeddings for all treebanks with training data, capping the maximum number of sentences per treebank to 800.
We then used Ward's method to perform a hierarchical cluster analysis.

We found that the most stable clusters were for closely related languages.
There was also a tendency for treebanks containing old languages (i.e., Ancient Greek, Gothic, Latin and Old Church Slavonic) to cluster together.
One reason for these languages parsing well together could be that several of the 7 treebanks come from the same annotation projects, four from PROIEL, and two from Perseus, containing consistently annotated and at least partially parallel data, e.g., from the Bible.

For the multi-treebank models, we performed preliminary experiments on development data investigating the effect of different groupings of languages. 
The main tendency we found was that it was better to use smaller groups of closely related languages rather than larger groups of slightly less related languages. 
For example, using multilingual models only for Galician-Portuguese and Spanish-Catalan was better than combining all Romance languages in a larger model, and combining Dutch-German-Afrikaans was better than also including English. 

A case where we use less related languages is for languages with very little training data (31 sentences or less), believing that it may be beneficial in this special case. We implemented this for Buryat, Uyghur and Kazakh, which are trained with Turkish, and Kurmanji, which is trained with Persian, even though these languages are not so closely related. 
For Armenian, which has only 50 training sentences, we could not find a close enough language, and instead train a single model on the available data.
For the four languages that are not in a multilingual cluster but have more than one available treebank, we use monolingual multi-treebank models (English, French, Italian and Korean). 

For the nine treebanks that have no training data we use different strategies:
\begin{itemize}[topsep=2pt,itemsep=1pt]
\item For Japanese Modern, we apply the mono-treebank Japanese GSD model.
\item For the four PUD treebanks, we apply the multi-treebank models trained using the other treebanks from that language, with the largest available treebank as proxy (except for Finnish, where we prefer Finnish TDT over FTB; cf.\ Section~\ref{sec:segmentation} and \citet{stymne+18acl}).
\item 
For Faroese, we apply the model for the Scandinavian languages, which are closely related, with Norwegian Nynorsk as proxy (cf.\ Section~\ref{sec:tagger}). 
In addition, we map the Faroese characters \{\'{I}\'{y}\'{u}ð\}, which do not occur in the other Scandinavian languages, to \{Iyud\}.
\item For Naija, an English-based creole, whose treebank according to the README file contains spoken language data, we train a special multilingual model on English EWT and the three small spoken treebanks for French, Norwegian, and Slovenian, and usd English EWT as proxy.\footnote{
We had found this combination to be useful in preliminary experiments where we tried to parse French Spoken without any French training data. }
\item 
For Thai and Breton, we create multilingual models trained with word and POS embeddings only (i.e., no character models or treebank embeddings) on Chinese and Irish, respectively. These models make use of multilingual word embeddings provided with Facebook's MUSE multilingual embeddings,\footnote{\url{https://github.com/facebookresearch/MUSE}} as described in more detail below.
\end{itemize}
For all multi-treebank models, we choose the model from the epoch that has the best mean LAS score among the treebanks that have available development data. 
This means that treebanks without development data rely on a model that is good for other languages in the group. 
In the cases of the mono-treebank Armenian and Irish models, where there is no development data, we choose the model from the final training epoch.
This also applies to the Breton model trained on Irish data.

\paragraph{Thai--Chinese} For the Thai model trained on Chinese, we were able to map Facebook's monolingual embeddings for each 
language to English using 
MUSE, thus creating multilingual Thai-Chinese embeddings.
We then trained a monolingual parser model using the mapped Chinese embeddings to initialize all word embeddings, and ensuring that these were not updated during training (unlike in the standard parser setup described in Section~\ref{sec:parsing}).
At test time, we look up all OOV word types, which are the great majority, in the mapped Thai embeddings first, otherwise assign them to a learned OOV vector.
Note that in this case, we had to increase the word embedding dimension in our parser to 300 to accomodate the larger Facebook embeddings.

\paragraph{Breton--Irish}
For Breton and Irish, the Facebook software does not come with the necessary resources to map these languages into English.
Here we instead created a small dictionary by using all available parallel data from OPUS (Ubuntu, KDE and Gnome, a total of 350K text snippets), and training a statistical machine translation model using Moses \cite{Moses07}.
From the lexical word-to-word correspondences created, we kept all cases where the translation probabilities in both directions were at least 0.4 and the words were not identical (in order to exclude a lot of English noise in the data), resulting in a word list of 6027 words.
We then trained monolingual embeddings for Breton using word2vec on Wikipedia data, and mapped them directly to Irish using MUSE.
A parser model was then trained, similarly to the Thai-Chinese case, using Irish embeddings as initialization, turning off updates to the word embeddings, and applying the mapped Breton embeddings at test time.

\renewcommand{\arraystretch}{1}
\renewcommand{\tabcolsep}{5pt}
\begin{table*}[tbp]
\begin{center}
\begin{scriptsize}
\scalebox{0.85}{
\begin{tabular}{|ll||r|r|r||r|r||r|r||ccc|}
\hline
\textsc{language}&\textsc{treebank}&\multicolumn{2}{c|}{\textsc{las}}&\multicolumn{1}{c||}{\textsc{mlas}}&\textsc{sents}&\textsc{words}&\textsc{upos}&\textsc{ufeats}&\multicolumn{1}{c|}{\textsc{segmentation}}&\textsc{tagging}&\multicolumn{1}{|c|}{\textsc{parsing}}\\
\hline
\textsc{arabic}&\textsc{padt}&\cellcolor{green!50} 73.54&73.54&\cellcolor{green!20} 61.04&\cellcolor{red!40} 68.06&\cellcolor{green!50} 96.19&\cellcolor{green!20} 90.70&\cellcolor{green!50} 88.25&~&~&~\\
\hline
\textsc{armenian}&\textsc{armtdp}&23.90&23.90&6.97&\cellcolor{red!40} 57.44&\cellcolor{red!20} 93.20&\cellcolor{green!50} 75.39&\cellcolor{green!20} 54.45&~&~&~\\
\hline
\textsc{basque}&\textsc{bdt}&78.12&78.12&\cellcolor{green!20} 67.67&\cellcolor{green!50} 100.00&\cellcolor{green!50} 100.00&\cellcolor{green!50} 96.05&\cellcolor{green!20} 92.50&~&~&~\\
\hline
\textsc{bulgarian}&\textsc{btb}&\cellcolor{green!20} 88.69&88.69&\cellcolor{green!20} 81.20&\cellcolor{green!50} 95.58&99.92&\cellcolor{green!50} 98.85&\cellcolor{green!20} 97.51&~&~&~\\
\hline
\textsc{breton}&\textsc{keb}&\cellcolor{green!50} 33.62&33.62&\cellcolor{green!50} 13.91&\cellcolor{green!20} 91.43&90.97&\cellcolor{green!50} 85.01&\cellcolor{green!50} 70.26&\multicolumn{1}{c}{\textsc{french gsd}}&\multicolumn{2}{|c|}{\textsc{special*}}\\
\hline
\textsc{chinese}&\textsc{gsd}&\cellcolor{green!50} 69.17&69.17&\cellcolor{green!50} 59.53&\cellcolor{green!50} 99.10&\cellcolor{green!50} 93.52&\cellcolor{green!50} 89.15&\cellcolor{green!50} 92.35&~&~&~\\
\hline
\textsc{greek}&\textsc{gdt}&\cellcolor{green!20} 86.39&86.39&\cellcolor{green!20} 72.29&\cellcolor{green!50} 91.92&\cellcolor{red!20} 99.69&\cellcolor{green!20} 97.26&\cellcolor{green!20} 93.65&~&~&~\\
\hline
\textsc{hebrew}&\textsc{htb}&\cellcolor{green!50} 67.72&67.72&\cellcolor{red!20} 44.19&\cellcolor{green!20} 100.00&\cellcolor{green!50} 90.98&\cellcolor{red!20} 80.26&79.49&~&~&~\\
\hline
\textsc{hungarian}&\textsc{szeged}&\cellcolor{green!20} 73.97&73.97&56.22&\cellcolor{red!40} 94.57&\cellcolor{red!40} 99.78&\cellcolor{green!20} 94.60&\cellcolor{red!40} 86.87&~&~&~\\
\hline
\textsc{indonesian}&\textsc{gsd}&\cellcolor{green!20} 78.15&78.15&\cellcolor{green!20} 67.90&\cellcolor{green!50} 93.47&\cellcolor{red!20} 99.99&\cellcolor{green!50} 93.70&\cellcolor{green!20} 95.83&~&~&~\\
\hline
\textsc{irish}&\textsc{idt}&\cellcolor{green!20} 68.14&68.14&\cellcolor{green!50} 41.72&\cellcolor{green!20} 94.90&\cellcolor{green!20} 99.60&\cellcolor{green!50} 91.55&\cellcolor{green!20} 81.78&~&~&~\\
\hline
\textsc{japanese}&\textsc{gsd}&\cellcolor{green!50} 79.97&79.97&\cellcolor{green!50} 65.47&\cellcolor{red!40} 94.92&\cellcolor{green!50} 93.32&\cellcolor{green!50} 91.73&\cellcolor{green!20} 91.66&~&~&~\\
\hline
\textsc{japanese}&\textsc{modern}&\cellcolor{green!50} 28.27&28.27&\cellcolor{green!50} 11.82&\cellcolor{red!20} 0.00&\cellcolor{green!50} 72.76&\cellcolor{green!50} 54.60&\cellcolor{green!50} 71.06&\multicolumn{3}{c|}{\textsc{japanese gsd}}\\
\hline
\textsc{latvian}&\textsc{lvtb}&\cellcolor{green!20} 76.97&76.97&\cellcolor{green!20} 63.90&\cellcolor{red!40} 96.97&\cellcolor{green!50} 99.67&\cellcolor{green!50} 94.95&\cellcolor{green!20} 91.73&~&~&~\\
\hline
\textsc{old french}&\textsc{srcmf}&\cellcolor{red!20} 78.71&78.71&69.82&\cellcolor{red!40} 59.15&100.00&\cellcolor{green!20} 95.48&\cellcolor{green!20} 97.26&~&~&~\\
\hline
\textsc{romanian}&\textsc{rrt}&\cellcolor{green!20} 84.33&84.33&\cellcolor{green!20} 76.00&\cellcolor{green!50} 95.81&\cellcolor{green!50} 99.74&\cellcolor{green!20} 97.46&\cellcolor{green!20} 97.25&~&~&~\\
\hline
\textsc{thai}&\textsc{pud}&\cellcolor{yellow!50} \sout{4.86}&\sout{4.86}&\cellcolor{yellow!50} \sout{2.22}&\cellcolor{green!50} 11.69&\cellcolor{green!50} 69.93&\cellcolor{yellow!50} \sout{33.75}&\cellcolor{green!50} 65.72&\multicolumn{3}{c|}{\textsc{special*}}\\
\hline
\textsc{vietnamese}&\textsc{vtb}&\cellcolor{green!20} 46.15&46.15&\cellcolor{green!20} 40.03&\cellcolor{red!40} 88.69&\cellcolor{green!20} 86.71&\cellcolor{green!20} 78.89&\cellcolor{green!20} 86.43&~&~&~\\
\hline
\textsc{afrikaans}&\textsc{afribooms}&\cellcolor{red!20} 79.47&78.89&\cellcolor{red!20} 66.35&\cellcolor{green!50} 99.65&\cellcolor{red!40} 99.37&96.28&95.39&~&~&~\\
\textsc{dutch}&\textsc{alpino}&83.58&81.73&\cellcolor{green!20} 71.11&\cellcolor{red!40} 89.04&\cellcolor{red!40} 99.62&\cellcolor{green!20} 95.78&\cellcolor{green!20} 95.89&~&~&~\\
~&\textsc{lassysmall}&\cellcolor{green!20} 82.25&79.59&\cellcolor{green!20} 70.88&73.62&\cellcolor{green!20} 99.87&\cellcolor{green!50} 96.18&\cellcolor{green!20} 95.85&~&~&~\\
\textsc{german}&\textsc{gsd}&75.48&75.15&\cellcolor{green!50} 53.67&\cellcolor{red!40} 79.36&\cellcolor{red!40} 99.37&\cellcolor{green!50} 94.02&\cellcolor{green!50} 88.13&~&~&~\\
\hline
\textsc{ancient greek}&\textsc{perseus}&65.17&62.95&\cellcolor{green!20} 44.31&\cellcolor{green!50} 98.93&\cellcolor{green!50} 99.97&\cellcolor{green!50} 92.40&\cellcolor{green!20} 90.12&~&~&~\\
~&\textsc{proiel}&72.24&71.58&\cellcolor{green!20} 54.98&\cellcolor{green!50} 51.17&\cellcolor{red!40} 99.99&\cellcolor{green!20} 97.05&\cellcolor{green!20} 91.04&~&~&~\\
\textsc{gothic}&\textsc{proiel}&63.40&60.58&49.79&\cellcolor{green!20} 31.97&\cellcolor{green!20} 100.00&\cellcolor{red!20} 93.43&\cellcolor{green!20} 88.60&~&~&~\\
\textsc{latin}&\textsc{ittb}&\cellcolor{green!20} 83.00&82.55&\cellcolor{green!20} 75.38&\cellcolor{green!50} 94.54&\cellcolor{green!50} 99.99&\cellcolor{green!50} 98.34&\cellcolor{green!20} 96.78&~&~&~\\
~&\textsc{perseus}&\cellcolor{green!20} 58.32&49.86&\cellcolor{green!20} 37.57&\cellcolor{green!20} 98.41&\cellcolor{green!20} 100.00&\cellcolor{green!20} 88.73&\cellcolor{green!20} 78.86&~&~&~\\
~&\textsc{proiel}&\cellcolor{red!20} 64.10&63.85&51.45&\cellcolor{green!50} 37.64&\cellcolor{green!50} 100.00&\cellcolor{green!20} 96.21&\cellcolor{green!20} 91.46&~&~&~\\
\textsc{old church slavonic}&\textsc{proiel}&\cellcolor{green!20} 70.44&70.31&\cellcolor{green!20} 58.31&\cellcolor{green!50} 44.56&\cellcolor{red!20} 99.99&\cellcolor{green!50} 95.76&\cellcolor{green!20} 88.91&~&~&~\\
\hline
\textsc{buryat}&\textsc{bdt}&\cellcolor{green!50} 17.96&8.45&\cellcolor{red!20} 1.26&\cellcolor{green!20} 93.18&\cellcolor{green!20} 99.04&\cellcolor{green!50} 50.83&\cellcolor{green!20} 40.63&\multicolumn{1}{c|}{\textsc{russian syntagrus}}&~&~\\
\cline{10-10}
\textsc{kazakh}&\textsc{ktb}&\cellcolor{green!50} 31.93&23.85&\cellcolor{green!50} 8.62&\cellcolor{green!50} 94.21&\cellcolor{green!50} 97.40&\cellcolor{green!50} 61.72&\cellcolor{green!20} 48.45&\multicolumn{1}{c|}{\textsc{russian syntagrus}}&~&~\\
\cline{10-10}
\textsc{turkish}&\textsc{imst}&\cellcolor{green!20} 61.34&61.77&\cellcolor{green!20} 51.23&\cellcolor{red!40} 96.63&\cellcolor{red!40} 97.80&\cellcolor{green!50} 93.72&\cellcolor{green!50} 90.42&~&~&~\\
\textsc{uyghur}&\textsc{udt}&\cellcolor{green!20} 62.94&62.38&\cellcolor{green!20} 42.54&\cellcolor{green!50} 83.47&\cellcolor{green!50} 99.69&\cellcolor{green!50} 89.19&\cellcolor{green!20} 87.00&~&~&~\\
\hline
\textsc{catalan}&\textsc{ancora}&88.94&88.68&\cellcolor{green!20} 81.39&\cellcolor{green!50} 99.35&\cellcolor{red!40} 99.79&\cellcolor{green!20} 98.38&\cellcolor{green!20} 97.90&~&~&~\\
\textsc{spanish}&\textsc{ancora}&\cellcolor{green!20} 88.79&88.65&\cellcolor{green!20} 81.75&\cellcolor{red!40} 97.97&99.92&\cellcolor{green!50} 98.69&\cellcolor{green!20} 98.23&~&~&~\\
\hline
\textsc{croatian}&\textsc{set}&\cellcolor{green!20} 84.62&84.13&\cellcolor{green!50} 70.53&\cellcolor{green!50} 96.97&\cellcolor{green!50} 99.93&\cellcolor{green!50} 97.93&\cellcolor{green!50} 91.70&~&~&~\\
\textsc{serbian}&\textsc{set}&\cellcolor{green!20} 86.99&85.14&\cellcolor{green!20} 75.54&\cellcolor{green!50} 93.07&\cellcolor{red!40} 99.94&\cellcolor{green!20} 97.61&\cellcolor{green!20} 93.70&~&~&~\\
\textsc{slovenian}&\textsc{ssj}&\cellcolor{green!20} 87.18&87.28&\cellcolor{green!50} 77.81&\cellcolor{green!50} 93.23&\cellcolor{green!50} 99.62&\cellcolor{green!50} 97.99&\cellcolor{green!50} 94.73&~&~&~\\
~&\textsc{sst}&\cellcolor{green!50} 56.06&53.27&\cellcolor{green!50} 41.22&\cellcolor{green!20} 23.98&100.00&\cellcolor{green!20} 93.18&\cellcolor{green!20} 84.75&~&~&~\\
\hline
\textsc{czech}&\textsc{cac}&\cellcolor{green!20} 89.49&88.94&\cellcolor{green!50} 82.25&\cellcolor{green!50} 100.00&\cellcolor{red!40} 99.94&\cellcolor{green!50} 99.17&\cellcolor{green!20} 95.84&~&~&~\\
~&\textsc{fictree}&\cellcolor{green!20} 89.76&87.78&\cellcolor{green!20} 80.63&\cellcolor{green!50} 98.72&\cellcolor{red!40} 99.85&\cellcolor{green!50} 98.42&\cellcolor{green!20} 95.52&~&~&~\\
~&\textsc{pdt}&88.15&88.09&\cellcolor{green!50} 82.39&92.29&\cellcolor{green!50} 99.96&\cellcolor{green!50} 99.07&\cellcolor{green!50} 96.89&~&~&~\\
\cline{10-12}
~&\textsc{pud}&\cellcolor{green!20} 84.36&83.35&\cellcolor{green!20} 74.46&\cellcolor{green!20} 96.29&\cellcolor{green!50} 99.62&\cellcolor{green!20} 97.02&\cellcolor{green!20} 93.66&\multicolumn{3}{c|}{\textsc{czech pdt}}\\
\cline{10-12}
\textsc{polish}&\textsc{lfg}&\cellcolor{green!20} 93.14&92.85&\cellcolor{green!50} 84.09&99.74&\cellcolor{green!50} 99.91&\cellcolor{green!50} 98.57&\cellcolor{green!50} 94.68&~&~&~\\
~&\textsc{sz}&\cellcolor{green!20} 89.80&88.48&\cellcolor{green!50} 77.28&\cellcolor{red!20} 98.91&\cellcolor{green!20} 99.94&\cellcolor{green!50} 97.95&\cellcolor{green!20} 91.82&~&~&~\\
\textsc{slovak}&\textsc{snk}&\cellcolor{green!50} 86.34&83.80&\cellcolor{green!50} 71.15&\cellcolor{green!50} 88.11&\cellcolor{red!20} 99.98&\cellcolor{green!50} 96.57&\cellcolor{green!50} 89.51&~&~&~\\
\cline{10-10}
\textsc{upper sorbian}&\textsc{ufal}&28.85&2.70&\cellcolor{red!20} 3.43&\cellcolor{green!20} 73.40&\cellcolor{red!20} 95.15&\cellcolor{red!40} 58.91&\cellcolor{red!20} 42.10&\multicolumn{1}{c|}{\textsc{spanish ancora}}&~&~\\
\hline
\textsc{danish}&\textsc{ddt}&80.08&79.68&\cellcolor{green!20} 71.19&\cellcolor{green!20} 90.10&\cellcolor{red!20} 99.85&\cellcolor{green!20} 97.14&\cellcolor{green!20} 97.03&~&~&~\\
\cline{10-12}
\textsc{faroese}&\textsc{oft}&\cellcolor{green!20} 41.69&39.94&\cellcolor{green!50} 0.70&\cellcolor{green!50} 95.32&\cellcolor{green!50} 99.25&\cellcolor{green!50} 65.54&\cellcolor{green!50} 34.56&\textsc{danish ddt}&\multicolumn{2}{|c|}{\textsc{norwegian nynorsk}}\\
\cline{10-12}
\textsc{norwegian}&\textsc{bokmaal}&\cellcolor{green!20} 88.30&87.68&\cellcolor{green!20} 81.68&\cellcolor{red!40} 95.13&\cellcolor{green!50} 99.84&\cellcolor{green!50} 98.04&\cellcolor{green!20} 97.18&~&~&~\\
~&\textsc{nynorsk}&\cellcolor{green!20} 87.40&86.23&\cellcolor{green!20} 79.42&\cellcolor{red!20} 92.09&\cellcolor{green!20} 99.94&\cellcolor{green!20} 97.57&\cellcolor{green!20} 96.88&~&~&~\\
~&\textsc{nynorsklia}&\cellcolor{green!20} 59.66&55.51&\cellcolor{green!20} 45.51&\cellcolor{green!20} 99.86&\cellcolor{green!20} 99.99&\cellcolor{green!20} 90.02&\cellcolor{green!20} 89.62&~&~&~\\
\textsc{swedish}&\textsc{lines}&\cellcolor{green!20} 80.53&78.33&\cellcolor{green!20} 65.38&\cellcolor{red!20} 85.17&\cellcolor{green!50} 99.99&\cellcolor{green!50} 96.64&\cellcolor{green!20} 89.54&~&~&~\\
\cline{10-12}
~&\textsc{pud}&\cellcolor{green!20} 78.15&75.52&\cellcolor{green!50} 49.73&\cellcolor{red!20} 91.57&\cellcolor{green!20} 98.78&\cellcolor{green!20} 93.12&\cellcolor{green!20} 78.53&\multicolumn{3}{c|}{\textsc{swedish talbanken}}\\
\cline{10-12}
~&\textsc{talbanken}&\cellcolor{green!20} 84.26&83.29&\cellcolor{green!20} 76.74&\cellcolor{green!50} 96.45&\cellcolor{green!50} 99.96&\cellcolor{green!20} 97.45&\cellcolor{green!20} 96.82&~&~&~\\
\hline
\textsc{english}&\textsc{ewt}&\cellcolor{green!20} 81.47&81.18&\cellcolor{green!20} 72.98&\cellcolor{red!20} 75.41&\cellcolor{green!20} 99.10&\cellcolor{green!50} 95.28&\cellcolor{green!20} 96.02&~&~&~\\
~&\textsc{gum}&\cellcolor{green!20} 81.28&79.23&\cellcolor{green!20} 69.62&\cellcolor{green!50} 81.16&\cellcolor{red!40} 99.71&\cellcolor{green!20} 94.67&\cellcolor{green!20} 95.80&~&~&~\\
~&\textsc{lines}&\cellcolor{green!20} 78.64&76.28&\cellcolor{green!20} 70.18&\cellcolor{green!20} 88.18&\cellcolor{green!20} 99.96&\cellcolor{green!20} 96.47&\cellcolor{green!20} 96.52&~&~&~\\
\cline{10-12}
~&\textsc{pud}&\cellcolor{green!20} 84.09&83.67&\cellcolor{green!20} 72.49&\cellcolor{green!50} 97.02&99.69&\cellcolor{green!20} 95.23&\cellcolor{green!20} 95.16&\multicolumn{3}{c|}{\textsc{english ewt}}\\
\hline
\textsc{estonian}&\textsc{edt}&\cellcolor{green!20} 81.09&81.47&\cellcolor{green!20} 74.11&\cellcolor{green!50} 92.16&\cellcolor{green!50} 99.96&\cellcolor{green!50} 97.16&\cellcolor{green!20} 95.80&~&~&~\\
\textsc{finnish}&\textsc{ftb}&\cellcolor{green!20} 84.19&83.12&\cellcolor{green!20} 76.40&\cellcolor{green!50} 87.91&\cellcolor{red!20} 99.98&\cellcolor{green!50} 96.30&\cellcolor{green!50} 96.73&~&~&~\\
\cline{10-12}
~&\textsc{pud}&\cellcolor{green!20} 86.48&86.48&\cellcolor{green!20} 80.52&\cellcolor{green!50} 92.95&\cellcolor{green!50} 99.69&\cellcolor{green!20} 97.59&\cellcolor{green!20} 96.84&\multicolumn{3}{c|}{\textsc{finnish tdt}}\\
\cline{10-12}
~&\textsc{tdt}&\cellcolor{green!20} 84.33&84.24&\cellcolor{green!20} 77.50&\cellcolor{green!50} 91.12&\cellcolor{green!50} 99.78&\cellcolor{green!50} 97.06&\cellcolor{green!20} 95.58&~&~&~\\
\cline{10-10}
\textsc{north saami}&\textsc{giella}&\cellcolor{green!20} 64.85&64.14&\cellcolor{green!20} 51.67&\cellcolor{green!20} 98.27&99.32&\cellcolor{green!20} 90.44&\cellcolor{green!20} 85.03&\multicolumn{1}{c|}{\textsc{german gsd}} &~&~\\
\hline
\textsc{french}&\textsc{gsd}&\cellcolor{green!20} 85.61&85.16&\cellcolor{green!20} 76.79&\cellcolor{green!50} 95.40&\cellcolor{green!50} 99.30&\cellcolor{green!50} 96.86&\cellcolor{green!20} 96.26&~&~&~\\
~&\textsc{sequoia}&\cellcolor{green!20} 87.39&86.26&\cellcolor{green!20} 79.97&\cellcolor{green!50} 87.33&\cellcolor{green!50} 99.44&\cellcolor{green!50} 97.92&\cellcolor{green!20} 97.47&~&~&~\\
~&\textsc{spoken}&\cellcolor{green!20} 71.26&69.44&\cellcolor{green!20} 60.12&\cellcolor{green!50} 23.54&100.00&\cellcolor{green!20} 95.51&100.00&~&~&~\\
\hline
\textsc{galician}&\textsc{ctg}&\cellcolor{red!20} 78.41&78.27&65.52&\cellcolor{red!20} 96.46&\cellcolor{red!40} 98.01&\cellcolor{red!40} 95.80&97.78&~&~&~\\
~&\textsc{treegal}&\cellcolor{green!20} 72.67&70.16&\cellcolor{green!50} 58.22&82.97&97.90&\cellcolor{green!20} 93.25&\cellcolor{green!20} 92.15&~&~&~\\
\textsc{portuguese}&\textsc{bosque}&\cellcolor{red!20} 84.41&84.27&\cellcolor{green!20} 71.76&\cellcolor{green!50} 90.89&\cellcolor{red!20} 99.00&\cellcolor{red!20} 95.90&\cellcolor{green!20} 95.41&~&~&~\\
\hline
\textsc{hindi}&\textsc{hdtb}&89.37&89.23&\cellcolor{green!20} 74.62&\cellcolor{red!20} 99.02&\cellcolor{green!20} 100.00&\cellcolor{green!50} 97.44&\cellcolor{green!20} 93.55&~&~&~\\
\textsc{urdu}&\textsc{udtb}&80.40&79.85&52.15&\cellcolor{green!20} 98.60&\cellcolor{green!20} 100.00&\cellcolor{green!20} 93.66&80.78&~&~&~\\
\hline
\textsc{italian}&\textsc{isdt}&\cellcolor{green!20} 89.43&89.37&\cellcolor{green!20} 81.17&\cellcolor{green!50} 99.38&\cellcolor{red!40} 99.55&\cellcolor{green!20} 97.79&\cellcolor{green!20} 97.36&~&~&~\\
~&\textsc{postwita}&\cellcolor{green!50} 76.75&76.46&\cellcolor{green!50} 66.46&\cellcolor{green!50} 54.00&\cellcolor{red!40} 99.04&\cellcolor{green!20} 95.61&\cellcolor{green!20} 95.63&~&~&~\\
\hline
\textsc{korean}&\textsc{gsd}&\cellcolor{green!20} 81.92&81.12&\cellcolor{green!20} 77.25&\cellcolor{green!50} 92.78&\cellcolor{green!50} 99.87&\cellcolor{green!20} 95.61&\cellcolor{green!20} 99.63&~&~&~\\
~&\textsc{kaist}&\cellcolor{green!20} 84.98&84.74&\cellcolor{green!20} 78.90&\cellcolor{green!20} 100.00&100.00&\cellcolor{green!20} 95.21&100.00&~&~&~\\
\hline
\textsc{kurmanji}&\textsc{mg}&\cellcolor{green!50} 29.54&7.61&\cellcolor{green!20} 5.77&\cellcolor{green!50} 90.85&\cellcolor{green!50} 96.97&\cellcolor{green!20} 61.33&\cellcolor{green!20} 48.26&\multicolumn{1}{c|}{\textsc{spanish ancora}}&~&~\\
\cline{10-10}
\textsc{persian}&\textsc{seraji}&83.39&83.22&\cellcolor{green!20} 76.97&\cellcolor{green!50} 99.50&99.60&\cellcolor{green!20} 96.79&\cellcolor{green!20} 97.02&~&~&~\\
\hline
\textsc{naija}&\textsc{nsc}&\cellcolor{green!20} 20.44&19.44&3.55&\cellcolor{red!20} 0.00&\cellcolor{green!50} 98.53&\cellcolor{green!50} 57.19&\cellcolor{red!20} 36.09&\multicolumn{2}{c|}{\textsc{english ewt}}&\multicolumn{1}{c|}{\textsc{special*}}\\
\hline
\textsc{russian}&\textsc{syntagrus}&89.00&89.39&\cellcolor{green!20} 81.01&\cellcolor{green!50} 98.79&99.61&\cellcolor{green!50} 98.59&\cellcolor{green!20} 94.89&~&~&~\\
~&\textsc{taiga}&\cellcolor{green!20} 65.49&59.32&\cellcolor{green!20} 46.07&\cellcolor{red!40} 66.40&97.81&\cellcolor{green!20} 89.32&\cellcolor{green!20} 82.15&~&~&~\\
\textsc{ukrainian}&\textsc{iu}&\cellcolor{green!20} 82.70&81.41&\cellcolor{red!20} 59.15&\cellcolor{red!40} 93.42&\cellcolor{green!50} 99.76&\cellcolor{green!50} 96.89&\cellcolor{red!20} 81.95&~&~&~\\
\hline
\hline
\textsc{all}&\cellcolor{yellow!50}\textsc{official}&\cellcolor{yellow!50} \sout{72.37}&\sout{70.71}&\cellcolor{yellow!50} \sout{59.20}&\cellcolor{green!20} 83.80&\cellcolor{green!50} 98.18&\cellcolor{yellow!50} \sout{90.91}&\cellcolor{green!20} 87.59&~&~&~\\
\hline
\textsc{all}&\cellcolor{green!50}\textsc{corrected}&\cellcolor{green!20} 72.31&70.65&\cellcolor{green!20} 59.17&\cellcolor{green!20} 83.80&\cellcolor{green!50} 98.18&\cellcolor{green!50} 90.50&\cellcolor{green!20} 87.59&~&~&~\\
\hline
\hline
\textsc{big}&~&\cellcolor{green!20} 80.25&79.61&\cellcolor{green!20} 68.81&\cellcolor{green!20} 87.23&\cellcolor{green!50} 99.10&\cellcolor{green!50} 95.59&\cellcolor{green!20} 93.65&~&~&~\\
\hline
\textsc{pud}&~&\cellcolor{green!20} 72.27&71.46&\cellcolor{green!50} 57.80&\cellcolor{green!20} 75.57&\cellcolor{green!50} 94.11&\cellcolor{green!50} 87.51&\cellcolor{green!50} 87.05&~&~&~\\
\hline
\textsc{small}&~&\cellcolor{green!20} 63.60&60.06&\cellcolor{green!50} 46.00&80.68&99.23&\cellcolor{green!50} 90.93&\cellcolor{green!20} 84.91&~&~&~\\
\hline
\hline
\textsc{low-resource}&\cellcolor{yellow!50}\textsc{official}&\cellcolor{yellow!50} \sout{25.87}&\sout{18.26}&\cellcolor{yellow!50} \sout{5.16}&\cellcolor{green!20} 67.50&\cellcolor{green!50} 93.38&\cellcolor{yellow!50} \sout{61.07}&\cellcolor{green!50} 48.95&~&~&~\\
\hline
\textsc{low-resource}&\cellcolor{green!50}\textsc{corrected}&\cellcolor{green!50} 25.33&17.72&\cellcolor{green!50} 4.91&\cellcolor{green!20} 67.50&\cellcolor{green!50} 93.38&\cellcolor{green!50} 57.32&\cellcolor{green!50} 48.95&~&~&~\\
\hline

\end{tabular}
}
\caption{Results for LAS (+ mono-treebank baseline), MLAS, sentence and word segmentation, UPOS tagging and morphological features (UFEATS). 
Treebanks sharing a parsing model grouped together;
substitute and proxy treebanks for segmentation, tagging, parsing far right
(\textsc{special} models detailed in the text).
Confidence intervals for coloring:
\colorbox{red!40}{$|$} 
$< \mu \!-\! \sigma <$
\colorbox{red!20}{$|$} 
$< \mu \!-\! 
\textsc{se} < \mu < \mu \!+\! 
\textsc{se} <$
\colorbox{green!20}{$|$}
$< \mu \!+\! \sigma <$
\colorbox{green!50}{$|$}.
}
\label{tab:results}
\end{scriptsize}
\end{center}
\end{table*}

\section{Results and Discussion}
\label{sec:results}

Table~\ref{tab:results} shows selected test results for the Upp\-sala system, including the two main metrics LAS and MLAS (plus a mono-treebank baseline for LAS),\footnote{Since our system does not predict lemmas, the third main metric BLEX is not very meaningful.} the sentence and word segmentation accuracy, and the accuracy of UPOS tagging and morphological features (UFEATS). To make the table more readable, we have added a simple color coding. Scores that are significantly higher/lower than the mean score of the 21 systems that successfully parsed all test sets are marked with two shades of green/red. The lighter shade marks differences that are outside the interval defined by the standard error of the mean ($\mu \pm 
\textsc{se},
\textsc{se} = \sigma / \sqrt{N}$) but within one standard deviation (std dev) from the mean. The darker shade marks differences that are more than one std dev above/below the mean ($\mu \pm \sigma$). Finally, scores that are no longer valid because of the Thai UPOS tagger are crossed out in yellow cells, and corrected scores are 
added where relevant.

Looking first at the LAS scores, we see that our results are significantly above the mean for all aggregate sets of treebanks (\textsc{all}, \textsc{big}, \textsc{pud}, \textsc{small}, \textsc{low-resource}) with an especially strong result for the low-resource group (even after setting the Thai score to 0.00).
If we look at specific languages, we do particularly well on low-resource languages like Breton, Buryat, Kazakh and Kurmanji, but also on languages like Arabic, Hebrew, Japanese and Chinese, where we benefit from having better word segmentation than most other systems. Our results are significantly worse than the mean only for Afrikaans AfriBooms, Old French SRCMF, Galician CTG, Latin PROIEL, and Portuguese Bosque. 
For Galician and Portuguese, this may be the effect of lower word segmentation and tagging accuracy. 

To find out whether our multi-treebank and multi-lingual models were in fact beneficial for parsing accuracy, we ran a post-evaluation experiment with one model per test set, each trained only on a single treebank. We refer to this as the mono-treebank baseline, and the LAS scores can be found in the second (uncolored) LAS column in Table~\ref{tab:results}. The results show that merging treebanks and languages did in fact improve parsing accuracy in a remarkably consistent fashion. For the 64 test sets that were parsed with a multi-treebank model, only four had a (marginally) higher score with the mono-treebank baseline model: Estonian EDT, Russian SynTagRus, Slovenian SSJ, and Turkish IMST. Looking at the aggregate sets, we see that, as expected, the pooling of resources helps most for \textsc{low-resource} (25.33 vs.\ 17.72) and \textsc{small} (63.60 vs.\ 60.06), but even for \textsc{big} there is some improvement (80.21 vs.\ 79.61). We find these results very encouraging, as they indicate that our treebank embedding method is a reliable method for pooling training data both within and across languages. It is also worth noting that this method is easy to use and does not require extra external resources used in most work on multilingual parsing, like multilingual word embeddings \cite{ammar16} or linguistic re-write rules \cite{aufrant+16} to achieve good results.

Turning to the MLAS scores, we see a very similar picture, but our results are relatively speaking stronger also for \textsc{pud} and \textsc{small}. There are a few striking reversals, where we do significantly better than the mean for LAS but significantly worse for MLAS, including Buryat BDT, Hebrew HTB and Ukrainian IU. Buryat and Ukrainian are languages for which we use a multilingual model for parsing, but not for UPOS tagging and morphological features, so it may be due to sparse data for tags and morphology, since these languages have very little training data. This is supported by the observation that low-resource languages in general have a larger drop from LAS to MLAS than other languages.

For sentence segmentation, the Uppsala system achieved the second best scores overall, and results are significantly above the mean for all aggregates except \textsc{small}, which perhaps indicates a sensitivity to data sparseness for the data-driven joint sentence and word segmenter (we see the same pattern for word segmentation). However, there is a much larger variance in the results than for the parsing scores, with altogether 23 treebanks having scores significantly below the mean
. 

For word segmentation, we obtained the best results overall, strongly outperforming the mean for all groups except \textsc{small}. We know from previous work \citep{shao18} that our word segmenter performs well on more challenging languages like Arabic, Hebrew, Japanese, and Chinese (although we were beaten by the Stanford team for the former two and by the HIT-SCIR team for the latter two). By contrast, it sometimes falls below the mean for the easier languages, but typically only by a very small fraction (for example 99.99 vs.\ 100.00 for 3 treebanks). Finally, it is worth noting that the maximum-matching segmenter developed specifically for Thai achieved a score of 69.93, which was more than 5 points better than any other system.

Our results for UPOS tagging indicate that this may be the strongest component of the system, although it is clearly helped by getting its input from a highly accurate word segmenter. The Upp\-sala system ranks first overall with scores more than one std dev above the mean for all aggregates. There is also much less variance than in the segmentation results, and scores are significantly below the mean only for five treebanks: Galician CTG, Gothic PROIEL, Hebrew HTB, Upper Sorbian UFAL, and Portuguese Bosque. For Galician and Upper Sorbian, the result can at least partly be explained by a lower-than-average word segmentation accuracy.

The results for morphological features are similar to the ones for UPOS tagging, with the best overall score but with less substantial improvements over the mean. The four treebanks where scores are significantly below the mean are all languages with little or no training data: Upper Sorbian UFAL, Hungarian Szeged, Naija NSC and Ukrainian IU. 

All in all, the 2018 edition of the Uppsala parser can be characterized as a system that is strong on segmentation (especially word segmentation) and prediction of UPOS tags and morphological features, and where the dependency parsing component performs well in low-resource scenarios thanks to the use of multi-treebank models, both within and across languages. 
For what it is worth, we also seem to have the highest ranking single-parser transition-based system in a task that is otherwise dominated by graph-based models, in particular variants of the winning Stanford system from 2017 \citep{dozat-conll-2017}.

\section{Extrinsic Parser Evaluation}
\label{sec:epe}

In addition to the official shared task evaluation, we also participated in the 2018 edition of the Extrinsic Parser Evaluation Initiative (EPE) \citep{epe18}, where parsers developed for the CoNLL 2018 shared task were evaluated with respect to their contribution to three downstream systems: biological event extraction, fine-grained opinion analysis, and negation resolution. The downstream systems are available for English only, and we participated with our English model trained on English EWT, English LinES and English GUM, using English EWT as the proxy. 

In the extrinsic evaluation, the Uppsala system ranked second for event extraction, first for opinion analysis, and 16th (out of 16 systems) for negation resolution. Our results for the first two tasks are better than expected, given that our system ranks in the middle with respect to intrinsic evaluation on English (9th for LAS, 6th for UPOS). By contrast, our performance is very low on the negation resolution task, which we suspect is due to the fact that our system only predicts universal part-of-speech tags (UPOS) and not the language specific PTB tags (XPOS), since the three systems that only predict UPOS are all ranked at the bottom of the list.

\section{Conclusion}
\label{sec:conc}

We have described the Uppsala submission to the CoNLL 2018 shared task, consisting of a segmenter that jointly extracts words and sentences from a raw text, a tagger that provides UPOS tags and morphological features, and a parser that builds a dependency tree given the words and tags of each sentence. For the parser we applied multi-treebank models both monolingually and multilingually, resulting in only 34 models for 82 treebanks as well as significant improvements in parsing accuracy especially for low-resource languages. We ranked 7th for the official LAS and MLAS scores, and first for the unofficial scores on word segmentation, UPOS tagging and morphological features.

\section*{Acknowledgments}

We are grateful to the shared task organizers and to Dan Zeman and Martin Potthast in particular, and we acknowledge the computational resources provided by CSC in Helsinki and Sigma2 in Oslo through NeIC-NLPL (www.nlpl.eu). Aaron Smith was supported by the Swedish Research Council.

\bibliography{expanded,main}

\begin{thebibliography}{}
\expandafter\ifx\csname natexlab\endcsname\relax\def\natexlab#1{#1}\fi

\bibitem[{Adi et~al.(2017)Adi, Kermany, Belinkov, Lavi, and
  Goldberg}]{adi17fine}
Yossi Adi, Einat Kermany, Yonatan Belinkov, Ofer Lavi, and Yoav Goldberg. 2017.
\newblock {Fine-grained Analysis of Sentence Embeddings Using Auxiliary
  Prediction Tasks}.
\newblock In {\em International Conference on Learning Representations\/}.

\bibitem[{Ammar et~al.(2016)Ammar, Mulcaire, Ballesteros, Dyer, and
  Smith}]{ammar16}
Waleed Ammar, George Mulcaire, Miguel Ballesteros, Chris Dyer, and Noah Smith.
  2016.
\newblock {Many Languages, One Parser}.
\newblock {\em Transactions of the Association for Computational Linguistics\/}
  4:431--444.

\bibitem[{Aufrant et~al.(2016)Aufrant, Wisniewski, and Yvon}]{aufrant+16}
Lauriane Aufrant, Guillaume Wisniewski, and Fran\c{c}ois Yvon. 2016.
\newblock {Zero-resource Dependency Parsing: Boosting Delexicalized
  Cross-lingual Transfer with Linguistic Knowledge}.
\newblock In {\em Proceedings of the 26th International Conference on
  Computational Linguistics (COLING)\/}. pages 119--130.

\bibitem[{Bahdanau et~al.(2014)Bahdanau, Cho, and Bengio}]{bahdanau2014neural}
Dzmitry Bahdanau, Kyunghyun Cho, and Yoshua Bengio. 2014.
\newblock {Neural Machine Translation by Jointly Learning to Align and
  Translate}.
\newblock {\em arXiv preprint arXiv:1409.0473\/} .

\bibitem[{Bohnet et~al.(2018)Bohnet, McDonald, Simoes, Andor, Pitler, and
  Maynez}]{bohnet18acl}
Bernd Bohnet, Ryan McDonald, Goncalo Simoes, Daniel Andor, Emily Pitler, and
  Joshua Maynez. 2018.
\newblock {Morphosyntactic Tagging with a {Meta-BiLSTM} Model over Context
  Sensitive Token Encodings}.
\newblock In {\em Proceedings of the 56th Annual Meeting of the Association for
  Computational Linguistics ({ACL})\/}.

\bibitem[{Bojanowski et~al.(2017)Bojanowski, Grave, Joulin, and
  Mikolov}]{bojanowski2016enriching}
Piotr Bojanowski, Edouard Grave, Armand Joulin, and Tomas Mikolov. 2017.
\newblock {Enriching Word Vectors with Subword Information}.
\newblock {\em Transactions of the Association for Computational Lingustics\/}
  5:135--146.

\bibitem[{Chen and Manning(2014)}]{chen14}
Danqi Chen and Christopher Manning. 2014.
\newblock {A Fast and Accurate Dependency Parser using Neural Networks}.
\newblock In {\em Proceedings of the Conference on Empirical Methods in Natural
  Language Processing ({EMNLP})\/}. pages 740--750.

\bibitem[{de~Lhoneux et~al.(2017{\natexlab{a}})de~Lhoneux, Shao, Basirat,
  Kiperwasser, Stymne, Goldberg, and Nivre}]{uu-conll17}
Miryam de~Lhoneux, Yan Shao, Ali Basirat, Eliyahu Kiperwasser, Sara Stymne,
  Yoav Goldberg, and Joakim Nivre. 2017{\natexlab{a}}.
\newblock {From Raw Text to Universal Dependencies -- Look, No Tags!}
\newblock In {\em {Proceedings of the CoNLL 2017 Shared Task: Multilingual
  Parsing from Raw Text to Universal Dependencies}\/}.

\bibitem[{de~Lhoneux et~al.(2017{\natexlab{b}})de~Lhoneux, Stymne, and
  Nivre}]{delhoneux17arc}
Miryam de~Lhoneux, Sara Stymne, and Joakim Nivre. 2017{\natexlab{b}}.
\newblock {Arc-Hybrid Non-Projective Dependency Parsing with a Static-Dynamic
  Oracle}.
\newblock In {\em {Proceedings of the 15th International Conference on Parsing
  Technologies}\/}. pages 99--104.

\bibitem[{Dozat et~al.(2017)Dozat, Qi, and Manning}]{dozat-conll-2017}
Timothy Dozat, Peng Qi, and Christopher~D. Manning. 2017.
\newblock {Stanford's Graph-based Neural Dependency Parser at the CoNLL 2017
  Shared Task}.
\newblock In {\em Proceedings of the CoNLL 2017 Shared Task: Multilingual
  Parsing from Raw Text to Universal Dependencies\/}. pages 20--30.

\bibitem[{Eryigit et~al.(2008)Eryigit, Nivre, and Oflazer}]{eryigit08cl}
G{\"u}lsen Eryigit, Joakim Nivre, and Kemal Oflazer. 2008.
\newblock {Dependency Parsing of {T}urkish}.
\newblock {\em Computational Linguistics\/} 34.

\bibitem[{Fares et~al.(2018)Fares, Oepen, Øvrelid, Bj{\"o}rne, and
  Johansson}]{epe18}
Murhaf Fares, Stephan Oepen, Lilja Øvrelid, Jari Bj{\"o}rne, and Richard
  Johansson. 2018.
\newblock The 2018 {S}hared {T}ask on {E}xtrinsic {P}arser {E}valuation. {O}n
  the downstream utility of {E}nglish universal dependency parsers.
\newblock In {\em Proceedings of the 22nd Conference on Computational Natural
  Language Learning (CoNLL)\/}.

\bibitem[{Gale and Church(1993)}]{gale1993program}
William~A Gale and Kenneth~W Church. 1993.
\newblock {A program for aligning sentences in bilingual corpora}.
\newblock {\em Computational linguistics\/} 19(1):75--102.

\bibitem[{Glorot and Bengio(2010)}]{glorot2010understanding}
Xavier Glorot and Yoshua Bengio. 2010.
\newblock {Understanding the difficulty of training deep feedforward neural
  networks.}
\newblock In {\em {Aistats}\/}. pages 249--256.

\bibitem[{Graves(2008)}]{graves2008bilstms}
Alex Graves. 2008.
\newblock {\em {{Supervised Sequence Labelling with Recurrent Neural
  Networks}}\/}.
\newblock Ph.D. thesis, Technical University Munich.

\bibitem[{Hochreiter and Schmidhuber(1997)}]{hochreiter1997long}
Sepp Hochreiter and J{\"u}rgen Schmidhuber. 1997.
\newblock {Long short-term memory}.
\newblock {\em Neural Computation\/} 9(8):1735--1780.

\bibitem[{Huang et~al.(2015)Huang, Xu, and Yu}]{huang2015bidirectional}
Zhiheng Huang, Wei Xu, and Kai Yu. 2015.
\newblock {Bidirectional {LSTM-CRF} models for sequence tagging}.
\newblock {\em arXiv preprint arXiv:1508.01991\/} .

\bibitem[{Hupkes et~al.(2018)Hupkes, Veldhoen, and
  Zuidema}]{hupkes2018visualisation}
Dieuwke Hupkes, Sara Veldhoen, and Willem Zuidema. 2018.
\newblock {Visualisation and `Diagnostic Classifiers' Reveal how Recurrent and
  Recursive Neural Networks Process Hierarchical Structure}.
\newblock {\em Journal of Artificial Intelligence Research\/} 61:907--926.

\bibitem[{Kapociute-Dzikiene et~al.(2013)Kapociute-Dzikiene, Nivre, and
  Krupavicius}]{kapociute13lithuanian}
Jurgita Kapociute-Dzikiene, Joakim Nivre, and Algis Krupavicius. 2013.
\newblock {Lithuanian dependency parsing with rich morphological features}.
\newblock In {\em {Proceedings of the fourth workshop on statistical parsing of
  morphologically-rich languages}\/}. pages 12--21.

\bibitem[{Kiperwasser and Goldberg(2016{\natexlab{a}})}]{kiperwasser16b}
Eliyahu Kiperwasser and Yoav Goldberg. 2016{\natexlab{a}}.
\newblock {Easy-First Dependency Parsing with Hierarchical Tree {LSTM}s}.
\newblock {\em Transactions of the Association for Computational Linguistics\/}
  4:445--461.

\bibitem[{Kiperwasser and Goldberg(2016{\natexlab{b}})}]{kiperwasser16}
Eliyahu Kiperwasser and Yoav Goldberg. 2016{\natexlab{b}}.
\newblock {Simple and Accurate Dependency Parsing Using Bidirectional {LSTM}
  Feature Representations}.
\newblock {\em Transactions of the Association for Computational Linguistics\/}
  4:313--327.

\bibitem[{Koehn et~al.(2007)Koehn, Hoang, Birch, Callison-Burch, Federico,
  Bertoldi, Cowan, Shen, Moran, Zens, Dyer, Bojar, Constantin, and
  Herbst}]{Moses07}
Philipp Koehn, Hieu Hoang, Alexandra Birch, Chris Callison-Burch, Marcello
  Federico, Nicola Bertoldi, Brooke Cowan, Wade Shen, Christine Moran, Richard
  Zens, Chris Dyer, Ondrej Bojar, Alexandra Constantin, and Evan Herbst. 2007.
\newblock {Moses: Open source toolkit for statistical machine translation}.
\newblock In {\em {Proceedings of the 45th Annual Meeting of the ACL, Demo and
  Poster Sessions}\/}. Prague, Czech Republic, pages 177--180.

\bibitem[{Mikolov et~al.(2013)Mikolov, Sutskever, Chen, Corrado, and
  Dean}]{mikolov2013distributed}
Tomas Mikolov, Ilya Sutskever, Kai Chen, Greg~S Corrado, and Jeff Dean. 2013.
\newblock {Distributed Representations of Words and Phrases and their
  Compositionality}.
\newblock In {\em {{Advances in Neural Information Processing Systems}}\/}.
  pages 3111--3119.

\bibitem[{Nivre(2008)}]{nivre2008algorithms}
Joakim Nivre. 2008.
\newblock {Algorithms for Deterministic Incremental Dependency Parsing}.
\newblock {\em Computational Linguistics\/} 34(4):513--553.

\bibitem[{Nivre(2009)}]{nivre09acl}
Joakim Nivre. 2009.
\newblock {Non-Projective Dependency Parsing in Expected Linear Time}.
\newblock In {\em Proceedings of the Joint Conference of the 47th Annual
  Meeting of the ACL and the 4th International Joint Conference on Natural
  Language Processing of the AFNLP (ACL-IJCNLP)\/}. pages 351--359.

\bibitem[{Nivre et~al.(2018)Nivre, Abrams, Agi{\'c} et~al.}]{ud22data}
Joakim Nivre, Mitchell Abrams, {\v Z}eljko Agi{\'c}, et~al. 2018.
\newblock \href{http://hdl.handle.net/11234/1-2837}{{{Universal Dependencies
  2.2}}}.
\newblock {LINDAT}/{CLARIN} digital library at the Institute of Formal and
  Applied Linguistics ({{\'U}FAL}), Faculty of Mathematics and Physics, Charles
  University.
\newblock
  \href{http://hdl.handle.net/11234/1-2837}{http://hdl.handle.net/11234/1-2837}.

\bibitem[{Potthast et~al.(2014)Potthast, Gollub, Rangel, Rosso, Stamatatos, and
  Stein}]{tira}
Martin Potthast, Tim Gollub, Francisco Rangel, Paolo Rosso, Efstathios
  Stamatatos, and Benno Stein. 2014.
\newblock {Improving the Reproducibility of {PAN}'s Shared Tasks: Plagiarism
  Detection, Author Identification, and Author Profiling}.
\newblock In Evangelos Kanoulas, Mihai Lupu, Paul Clough, Mark Sanderson, Mark
  Hall, Allan Hanbury, and Elaine Toms, editors, {\em {Information Access
  Evaluation. Multilinguality, Multimodality, and Visualization. 5th
  International Conference of the {CLEF} Initiative ({CLEF} 14)}\/}. pages
  268--299.

\bibitem[{Shao et~al.(2018)Shao, Hardmeier, and Nivre}]{shao18}
Yan Shao, Christian Hardmeier, and Joakim Nivre. 2018.
\newblock {Universal Word Segmentation: Implementation and Interpretation}.
\newblock {\em Transactions of the Association for Computational Linguistics\/}
  6:421--435.

\bibitem[{Shao et~al.(2017)Shao, Hardmeier, Tiedemann, and Nivre}]{shao17}
Yan Shao, Christian Hardmeier, J{\"o}rg Tiedemann, and Joakim Nivre. 2017.
\newblock {Character-Based Joint Segmentation and {POS} Tagging for {C}hinese
  using Bidirectional {RNN-CRF}}.
\newblock In {\em {The 8th International Joint Conference on Natural Language
  Processing}\/}. pages 173--183.

\bibitem[{Smith et~al.(2018)Smith, de~Lhoneux, Stymne, and Nivre}]{smith18}
Aaron Smith, Miryam de~Lhoneux, Sara Stymne, and Joakim Nivre. 2018.
\newblock {An Investigation of the Interactions Between Pre-Trained Word
  Embeddings, Character Models and POS Tags in Dependency Parsing}.
\newblock In {\em {Proceedings of the 2018 Conference on Empirical Methods in
  Natural Language Processing}\/}.

\bibitem[{Stymne et~al.(2018)Stymne, de~Lhoneux, Smith, and
  Nivre}]{stymne+18acl}
Sara Stymne, Miryam de~Lhoneux, Aaron Smith, and Joakim Nivre. 2018.
\newblock {Parser Training with Heterogeneous Treebanks}.
\newblock In {\em {Proceedings of the 56th Annual Meeting of the ACL, Short
  papers}\/}. pages 619--625.

\bibitem[{Tiedemann(2012)}]{tiedemann2012OPUS}
J\"{o}rg Tiedemann. 2012.
\newblock {Parallel Data, Tools and Interfaces in {OPUS}}.
\newblock In Nicoletta Calzolari~(Conference Chair), Khalid Choukri, Thierry
  Declerck, Mehmet~Ugur Dogan, Bente Maegaard, Joseph Mariani, Jan Odijk, and
  Stelios Piperidis, editors, {\em {Proceedings of the Eight International
  Conference on Language Resources and Evaluation (LREC'12)}\/}. European
  Language Resources Association (ELRA).

\bibitem[{Wong and Chan(1996)}]{wong96}
Pak-Kwong Wong and Chorkin Chan. 1996.
\newblock {Chinese Word Segmentation Based on Maximum Matching and Word Binding
  Force}.
\newblock In {\em {Proceedings of the 16th International Conference on
  Computational Linguistics}\/}. pages 200--203.

\bibitem[{Zeman et~al.(2018)Zeman, Haji{\v{c}}, Popel, Potthast, Straka,
  Ginter, Nivre, and Petrov}]{conll2018}
Daniel Zeman, Jan Haji{\v{c}}, Martin Popel, Martin Potthast, Milan Straka,
  Filip Ginter, Joakim Nivre, and Slav Petrov. 2018.
\newblock {CoNLL 2018 Shared Task: Multilingual Parsing from Raw Text to
  Universal Dependencies}.
\newblock In {\em Proceedings of the CoNLL 2018 Shared Task: Multilingual
  Parsing from Raw Text to Universal Dependencies\/}.

\bibitem[{Zeman et~al.(2017)Zeman, Popel, Straka, Haji{\v{c}}, Nivre, Ginter,
  Luotolahti, Pyysalo, Petrov, Potthast, Tyers, Badmaeva, G{\"{o}}k{\i}rmak,
  Nedoluzhko, Cinkov{\'{a}}, Haji{\v{c}}~jr., Hlav{\'{a}}{\v{c}}ov{\'{a}},
  Kettnerov{\'{a}}, Ure{\v{s}}ov{\'{a}}, Kanerva, Ojala, Missil{\"{a}},
  Manning, Schuster, Reddy, Taji, Habash, Leung, de~Marneffe, Sanguinetti,
  Simi, Kanayama, de~Paiva, Droganova, Mart{\'{\i}}nez~Alonso, Uszkoreit,
  Macketanz, Burchardt, Harris, Marheinecke, Rehm, Kayadelen, Attia, Elkahky,
  Yu, Pitler, Lertpradit, Mandl, Kirchner, Fernandez~Alcalde, Strnadova,
  Banerjee, Manurung, Stella, Shimada, Kwak, Mendon{\c{c}}a, Lando, Nitisaroj,
  and Li}]{udst:overview}
Daniel Zeman, Martin Popel, Milan Straka, Jan Haji{\v{c}}, Joakim Nivre, Filip
  Ginter, Juhani Luotolahti, Sampo Pyysalo, Slav Petrov, Martin Potthast,
  Francis Tyers, Elena Badmaeva, Memduh G{\"{o}}k{\i}rmak, Anna Nedoluzhko,
  Silvie Cinkov{\'{a}}, Jan Haji{\v{c}}~jr., Jaroslava
  Hlav{\'{a}}{\v{c}}ov{\'{a}}, V{\'{a}}clava Kettnerov{\'{a}}, Zde{\v{n}}ka
  Ure{\v{s}}ov{\'{a}}, Jenna Kanerva, Stina Ojala, Anna Missil{\"{a}},
  Christopher Manning, Sebastian Schuster, Siva Reddy, Dima Taji, Nizar Habash,
  Herman Leung, Marie-Catherine de~Marneffe, Manuela Sanguinetti, Maria Simi,
  Hiroshi Kanayama, Valeria de~Paiva, Kira Droganova, H{\v{e}}ctor
  Mart{\'{\i}}nez~Alonso, Hans Uszkoreit, Vivien Macketanz, Aljoscha Burchardt,
  Kim Harris, Katrin Marheinecke, Georg Rehm, Tolga Kayadelen, Mohammed Attia,
  Ali Elkahky, Zhuoran Yu, Emily Pitler, Saran Lertpradit, Michael Mandl, Jesse
  Kirchner, Hector Fernandez~Alcalde, Jana Strnadova, Esha Banerjee, Ruli
  Manurung, Antonio Stella, Atsuko Shimada, Sookyoung Kwak, Gustavo
  Mendon{\c{c}}a, Tatiana Lando, Rattima Nitisaroj, and Josie Li. 2017.
\newblock {{CoNLL 2017 Shared Task: Multilingual Parsing from Raw Text to
  Universal Dependencies}}.
\newblock In {\em {{Proceedings of the CoNLL 2017 Shared Task: Multilingual
  Parsing from Raw Text to Universal Dependencies}}\/}.

\end{thebibliography}
\bibliographystyle{acl_natbib}

\end{document}